\documentclass[conference]{IEEEtran}
\IEEEoverridecommandlockouts

\usepackage{float,epsfig,enumerate,bm,amsmath,color,soul,amssymb}
\usepackage{algorithm}
\usepackage{algorithmic}
\usepackage{graphicx}
\usepackage{caption}
\usepackage[labelformat=simple]{subcaption}

\usepackage{caption}
\usepackage{ulem}
\usepackage{cite}
\usepackage{amsthm}
\theoremstyle{plain}

\def\BibTeX{{\rm B\kern-.05em{\sc i\kern-.025em b}\kern-.08em
    T\kern-.1667em\lower.7ex\hbox{E}\kern-.125emX}}

\begin{document}

\title{A Semantic Communication-Based Workload-Adjustable Transceiver for Wireless AI-Generated Content (AIGC) Delivery}

\author{
	\IEEEauthorblockN{
		Runze Cheng\IEEEauthorrefmark{1},
		Yao Sun\IEEEauthorrefmark{1},
		Lan Zhang\IEEEauthorrefmark{2},
		Lei Feng\IEEEauthorrefmark{3},
		Lei Zhang\IEEEauthorrefmark{1}
		and Muhammad Ali Imran\IEEEauthorrefmark{1}}
	\IEEEauthorblockA{\IEEEauthorrefmark{1}James Watt School of Engineering, University of Glasgow, Glasgow, UK. \\
	\IEEEauthorrefmark{2}Department of Electrical and Computer Engineering, Clemson University, South Carolina, USA. \\
	\IEEEauthorrefmark{3}Beijing University of Posts and Telecommunications, Beijing, China.\\ Email: Yao.Sun@glasgow.ac.uk} 
    \thanks{This paper is partially funded by UK Department for Science, Innovation \& Technology (DSIT) Towards Ubiquitous 3D Open Resilient Network (TUDOR) Project.}
}
%

\maketitle

\begin{abstract}
	With the significant advances in generative AI (GAI) and the proliferation of mobile devices, providing high-quality AI-generated content (AIGC) services via wireless networks is becoming the future direction. However, the primary challenges of AIGC service delivery in wireless networks lie in unstable channels, limited bandwidth resources, and unevenly distributed computational resources. In this paper, we employ semantic communication (SemCom) in diffusion-based GAI models to propose a \underline{r}esource-aware w\underline{o}rkload-adj\underline{u}stable \underline{t}ransceiv\underline{e}r (ROUTE) for AIGC delivery in dynamic wireless networks. Specifically, to relieve the communication resource bottleneck, SemCom is utilized to prioritize semantic information of the generated content. Then, to improve computational resource utilization in both edge and local and reduce AIGC semantic distortion in transmission, modified diffusion-based models are applied to adjust the computing workload and semantic density in cooperative content generation. Simulations verify the superiority of our proposed ROUTE in terms of latency and content quality compared to conventional AIGC approaches.
\end{abstract}

%

\IEEEpeerreviewmaketitle

\section{Introduction}

Generative AI (GAI) has attracted a vast user base with its remarkable strides in analyzing various forms of media and creating high-quality AI-generated content (AIGC)~\cite{1}. To adapt to the traffic surge in wireless networks and provide inclusive and ubiquitous AIGC services, many AI companies, such as OpenAI and Meta, are actively strategizing and positioning themselves in the realm of wireless device-oriented AIGC applications~\cite{2}.

However, existing wireless AIGC services are delivered from cloud servers, which have been facing surging service demands from the tremendous globally distributed user base, sometimes becoming overloaded. In this context, some studies propose to offload content generation tasks from cloud servers to edge servers or local devices~\cite{2,3}. These frameworks reduce the pressure of cloud servers and improve the utilization of edge or local computational resources. Nevertheless, to cope with potentially overwhelming service requests for edge servers, communication resource constraint remains a common bottleneck~\cite{3}. In parallel, although some lightweight GAI models have been examined to be valid for locally generating content on local mobile devices~\cite{4}, the generated content quality and content-generating latency are difficult to ensure. Additionally, the fluctuating channel conditions and dynamic availability of resources (like bandwidth and computational power) in wireless networks, complicate the situation. Fixed GAI models deployed solely on edge servers or local devices struggle to adjust content quality and service latency as per resource availability and channel conditions. Therefore, a new wireless AIGC delivery framework is required to provide low-latency, high-quality AIGC services under dynamic channel conditions with limited bandwidth resources and unevenly distributed computational resources.

Considering the aforementioned constraints, exploiting the diffusion-based GAI and semantic communications (SemCom) in the AIGC delivery framework emerges as a potential solution. Diffusion-based GAI, by learning intricate patterns and structures of diverse media and iteratively reconstructing semantic information from noise-infused data, has brought a significant improvement to the generated content quality for many AIGC applications~\cite{5}. By controlling the number of denoising steps in the diffusion-based GAI model, both semantic density and computational workload can be adjusted to avoid high semantic noise ratios in harsh channel conditions and prevent high service latency caused by limited computational resources, respectively~\cite{1}. Meanwhile, SemCom prioritizes the accurate conveyance of the meaning implied in source messages. Empowered by SemCom, the semantic information of AIGC can be transmitted instead of the entire bits of content, and the pressure of communication resources, especially bandwidth, can be relieved with a slim data size~\cite{6}. In addition, its receiver is capable of accurately interpreting semantic information from transmitted bits, with the potential to further minimize semantic distortion caused by channel noise, enabling reliable AIGC services even over challenging wireless channels~\cite{7}. More importantly, the diffusion-based GAI and SemCom generally employ similar encoder-decoder components with overlapping neural network layers, which is natural to integrate without the obvious increase in system complexity.

In this paper, we modify the diffusion-based models in response to channel-caused semantic noise and employ SemCom in the diffusion-based GAI models to propose a \underline{r}esource-aware w\underline{o}rkload-adj\underline{u}stable \underline{t}ransceiv\underline{e}r (ROUTE). In ROUTE, semantic information of AIGC is first generated at the transmitter, then delivered using SemCom, and finally fine-tuned locally at the receiver. Especially, the computing workloads and semantic density can be adjusted by adapting the number of denoising steps at both transmitter and receiver based on resource availability and channel quality. This allows for cooperative, high-quality, low-latency image generation between the edge server and local device. Simulation results demonstrate the effectiveness and robustness of our proposed ROUTE under dynamic resource availability and varying channel quality and verify the performance gain in terms of latency and image quality by comparing it with other wireless AIGC frameworks.\footnote{The corresponding code can be found at https://github.com/RunzeCheng/ ROUTE-transceiver.git.}

\section{Resource-Aware Workload-Adjustable Transceiver Design}

In this work, we specifically delve into image-based AIGC service delivering over wireless networks, and assuming the generation of images is from textual descriptions, i.e., text-to-image (T2I).

\subsection{ROUTE Framework}

Let us consider a downlink image generation and transmission scenario. The ROUTE consists of three stages, i.e., \textit{cloud pre-training} for remote model training, \textit{edge encoding} for semantic generation and transmission, and \textit{local decoding} for semantic fine-tuning and image reconstruction.

\begin{figure*}
	\centering
	\includegraphics[width=0.88\linewidth]{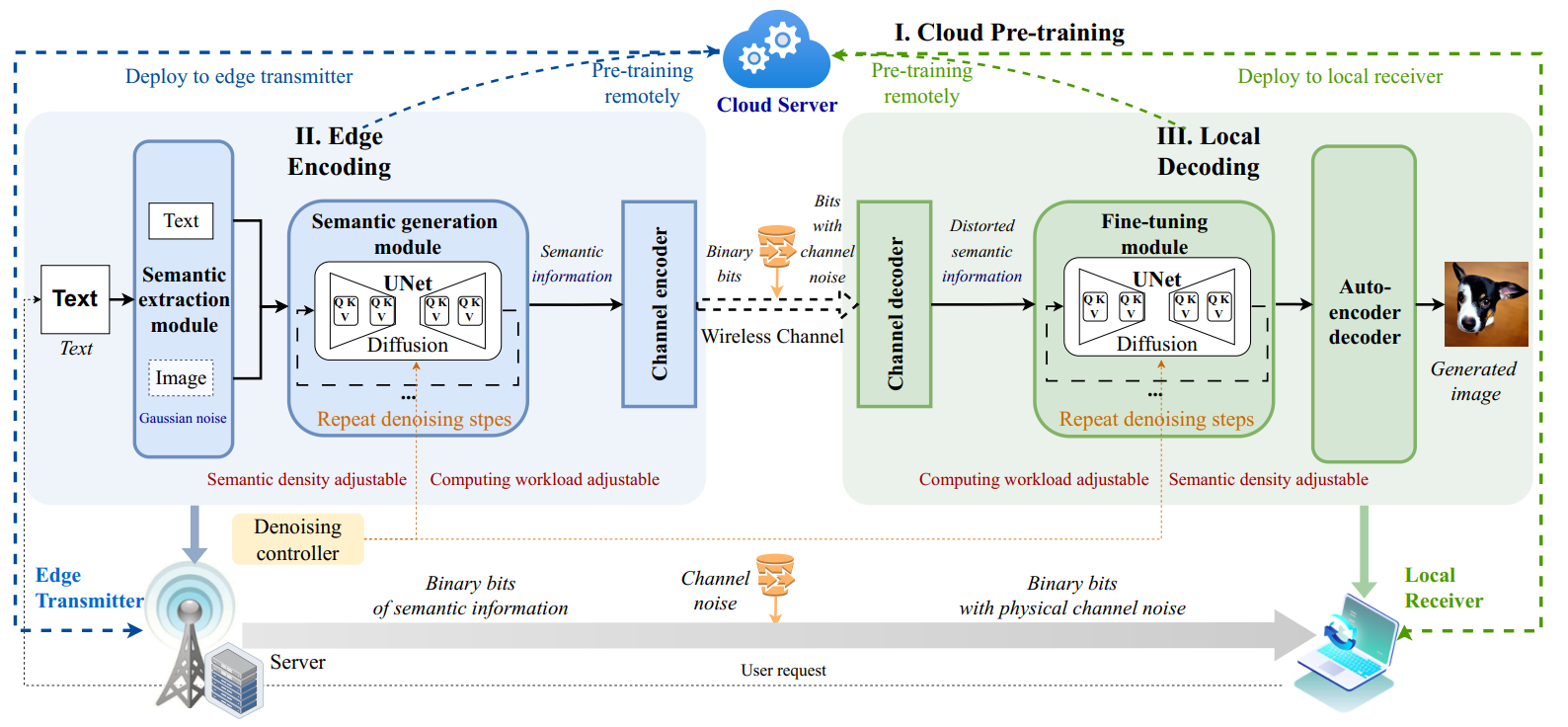}
	\caption{The framework of proposed ROUTE.}
	\label{fig:1}
\end{figure*}

\textit{Cloud pre-training:} A cloud server is utilized to conduct the diffusion-based T2I model training processes. Typically, high-quality T2I models require millions of parameters with a total size of 1-10 GB, which exposes stringent requirements for specialized data processing hardware during training. The cloud server as a centralized infrastructure with ample computing power, storage, database, etc. matches well to these demands. As shown in Fig.~\ref{fig:1}, the cloud server retains the copy of all the models within both the edge transmitter and local receiver. Since the large models within the transceiver generally perform stable after the initial offline pre-training, further model updates are unnecessary to be frequent. Even when periodic updates for the models are required, modularization of these models allows each update to target specific modules within ROUTE, as shown in Fig.~\ref{fig:1}, thereby leading to less data transmission.

\textit{Edge encoding:} As shown in Fig.~\ref{fig:1}, the edge transmitter consists of a semantic extraction module, a semantic generation module, and a channel encoder. The semantic extraction module comprises two independent networks, responsible for extracting semantic information from text and image, respectively. As the core of the edge transmitter, the semantic generation module uses a diffusion model to predict and gradually denoise data starting from pure noise according to the provided textual description. Additionally, the channel encoder in Fig.~\ref{fig:1} converts semantic information into binary bits for transmission as per channel conditions.

\textit{Local decoding:} With advancements in mobile device computing power, many user devices are now capable of executing information fine-tuning and recovery processes. In ROUTE, the T2I decoder of a local receiver in Fig.~\ref{fig:1} is composed of a channel decoder, a semantic fine-tuning module, and an autoencoder decoder module. Specifically, the channel decoder converts the binary bits into semantic information. Given the limited computing power of local devices, a lightweight diffusion model is embedded within the semantic fine-tuning module. After a rapid fine-tuning process, semantic noise can be eliminated, and then the autoencoder decoder module restructures semantic information into high-resolution images with text guidance.

\subsection{Encoder Design at Edge Transmitter}

As mentioned, the semantic extraction module, semantic generation module, and channel encoder are deployed in the edge transmitter, as shown in the left part of Fig.~\ref{fig:1}.

\textit{1) Semantic extraction module:} Let $\mathbf{s}_{\scriptscriptstyle t\!e\!x\!t}$ and $\mathbf{s}$ denote the input text and image, respectively. The two modalities are extracted as semantic information $\mathbf{z}_{\scriptscriptstyle t\!e\!x\!t}=\mathcal{E}\left(\mathbf{s}_{\scriptscriptstyle t\!e\!x\!t};\boldsymbol{\varphi}_{\scriptscriptstyle t\!e\!x\!t}\right)$ and $\mathbf{z}=\mathcal{E}\left(\mathbf{s};\boldsymbol{\varphi}\right)$, where $\mathcal{E}\left(\cdot;\boldsymbol{\varphi}_{\scriptscriptstyle t\!e\!x\!t}\right)$ and $\mathcal{E}\left(\cdot;\boldsymbol{\varphi}\right)$ are the contrastive language-image pre-training (CLIP)-based~\cite{8} text encoder and an attention-based variational autoencoder (VAE)-based~\cite{9} image encoder with learnable parameters $\boldsymbol{\varphi}_{\scriptscriptstyle t\!e\!x\!t}$ and $\boldsymbol{\varphi}$, respectively. Here, text semantic information is a sequence, and image semantic information is the latent space tensor generated from multiple sets of means and standard deviations of Gaussian-like distribution. Note that the semantic information of real images is only used during the model training stage and is not directly involved in the T2I generation.

\textit{2) Semantic-generation module:}

\begin{figure}
	\centering
	\includegraphics[width=0.84\linewidth]{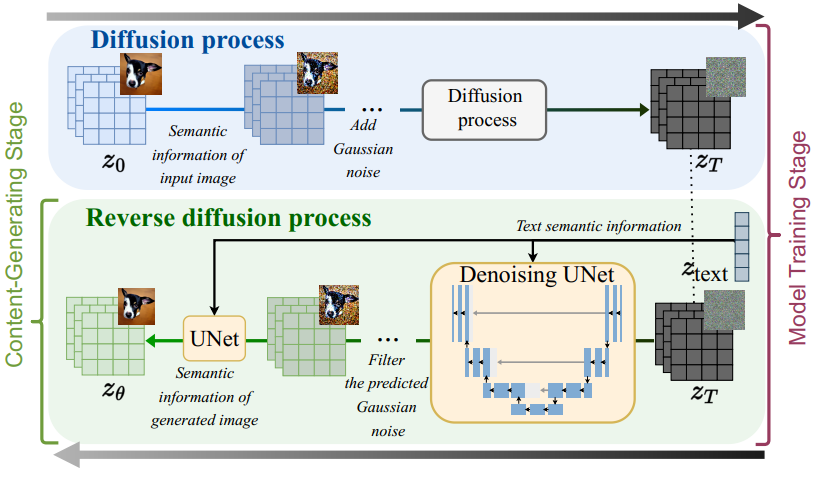}
	\caption{The diffusion and reverse diffusion processes.}
	\label{fig:2}
\end{figure}

The semantic generation module in Fig.~\ref{fig:1} is based on the denoising diffusion probabilistic model (DDPM)~\cite{5}. The core idea of this diffusion-based model is that if semantic information can be gradually polluted into pure Gaussian noise, then vice, pure Gaussian noise can be denoised into semantic information. In this context, it consists of two processes, i.e., a diffusion process and a reverse diffusion process. As shown in Fig.~\ref{fig:2}, the diffusion process $q(\mathbf{z}_t|\mathbf{z}_{t-1})$ is a fixed (or predefined) forward diffusion process, in which a scheduler gradually adds Gaussian noise at each time step $t\in[0,T]$, until the initial semantic information of an image $\mathbf{z}_{0}$ becomes pure noise $\mathbf{z}_T$. The forward diffusion process is
\begin{equation} \label{eq:1}
	\mathbf{z}_t=\sqrt{1-\beta_t}\mathbf{z}_{t-1}+\sqrt{\beta_t}\boldsymbol{\epsilon},
\end{equation}

\noindent where $0<\beta_1<\beta_2<\cdots<\beta_{\scriptscriptstyle T}<1$ is a known variance schedule with time-related constants, and $\boldsymbol{\epsilon}\sim\mathcal{N}(0,\boldsymbol{I})$ is the added Gaussian noise with an identity matrix $I.$ Meanwhile, the reverse diffusion process in Fig.~\ref{fig:2} is a denoising process, represented as $p(\mathbf{z}_{t-1}|\mathbf{z}_t).$ In this process, a neural network, i.e., UNet~\cite{9}, is trained to learn the conditional probability distribution $p_{\boldsymbol{\theta}}(\mathbf{z}_{t-1}|\mathbf{z}_t,\mathbf{z}_{\scriptscriptstyle t\!e\!x\!t})$ and gradually denoise the semantic information according to text semantic $\mathbf{z}_{\scriptscriptstyle t\!e\!x\!t}$ starting from pure noise until it ends up with the semantic information of an actual image. Here, $\boldsymbol{\theta}$ is the learnable parameters of the UNet, and text guidance $\mathbf{z}_{\scriptscriptstyle t\!e\!x\!t}$ is implemented by concatenating the text embedding to the key-value pairs of each self-attention layer in the UNet~\cite{9}. Accordingly, the reverse diffusion process is
\begin{equation} \label{eq:2}
	\mathbf{z}_{t-1}=\frac{1}{\sqrt{\alpha_t}}\left(\mathbf{z}_t-\frac{1-\alpha_t}{\sqrt{1-\dot{\alpha_t}}}\boldsymbol{\epsilon}_{\boldsymbol{\theta}}\left(\mathbf{z}_t,t,\mathbf{z}_{\scriptscriptstyle t\!e\!x\!t}\right)\right)+\bar{\sigma}_t\tilde{\boldsymbol{\epsilon}},
\end{equation}

\noindent where $\boldsymbol{\epsilon}_{\boldsymbol{\theta}}$ is the UNet predicted noise, $\alpha_t=1-\beta_t,t\in$ $\{1,\cdots,T\}$ is a known constant of step $t$, $\dot{\alpha}_t=\prod_{i=1}^{t}\alpha_i$ is a cumulative product of $\alpha_t$, $\bar{\sigma}_t$ is also a time dependent constant of step $t$,and $\tilde{\boldsymbol{\epsilon}}\sim\mathcal{N}(0,\boldsymbol{I})$ is the random normal Gaussian noise.

The semantic generation module operates differently in the pre-training stage and the content-generation (or edge encoding) stage, as shown in Fig.~\ref{fig:2}. In pre-training, it learns the noise distribution through both forward and reverse diffusion processes, while in content generation, only the reverse diffusion process is used. We introduce the content generating stage here, whereas the model training will be elaborated in Section III-A. In the content-generating stage, the image semantic information is generated by denoising randomly generated pure noisy data $\mathbf{z}_T$ as per text guidance $\mathbf{z}_{\boldsymbol{\theta}}=\mathcal{F}_1(\mathbf{z}_{T},T,\mathbf{z}_{\scriptscriptstyle t\!e\!x\!t};\boldsymbol{\theta})$, where $T$ is the denoising step number and $\mathcal{F}_1(\cdot;\boldsymbol{\theta})$ is the UNet with learnable parameter $\boldsymbol{\theta}$.

\textit{3) Channel encoder module:} As very few bits are required for text semantic information transmission, let us merely focus on the transmission of image semantic information~\cite{10}. The transmitted signal is encoded as $\mathbf{Y}=\mathcal{C}(\mathbf{z}_{\boldsymbol{\theta}})$ via the channel encoder, where $\mathbf{Y}$ is the analog signal.

After passing through the physical channel, the received signal is expressed as ${\mathbf{Y}}^{\prime}={\mathbf{H}\mathbf{Y}}+\mathbf{N}$, where $\mathbf{H}$ is the channel gain matrix and $\mathbf{N}$ represents the noise. This paper considers the additive white Gaussian noise (AWGN) channel.\footnote{The used diffusion model is Gaussian noise-based hot diffusion, which is designed to denoise Gaussian distributed noise. For filtering other types of noise, like Rayleigh noise, it is necessary to specifically pre-train a cold diffusion-based module~\cite{11}.}

\subsection{Decoder Design at Local Receiver}

Then, we present the T2I decoder of the local receiver, which consists of the channel decoder module, fine-tuning module, and autoencoder decoder module, as shown in the right part of Fig.~\ref{fig:1}.

\textit{1) Channel decoder module:} The received binary bits are decoded by traditional channel decoder as semantic information $\mathbf{z}^\prime=\mathcal{C}^{-1}(\mathbf{Y}^{\prime}),\mathbf{z}^{\prime}\approx\mathbf{z}+\boldsymbol{\epsilon}^{\mathcal{C}}$, where $\boldsymbol{\epsilon}^\mathcal{C}\sim\mathcal{N}(0,\sigma^2\boldsymbol{I})$ is the semantic noise with a variance matrix $\sigma^2\boldsymbol{I}$ caused by the physical channel noise $\mathbf{N}$.

\textit{2) Fine-tuning module:} The fine-tuning module employs a modified diffusion-based model, especially for denoising semantic noise caused by the channel. The reverse diffusion process of fine-tuning module is denoted as $\mathbf{z}_{\bar{\boldsymbol{\theta}}}=\mathcal{F}_2(\mathbf{z}^{\prime},\bar{T},\mathbf{z}_{\scriptscriptstyle t\!e\!x\!t};\bar{\boldsymbol{\theta}})$, where $\mathcal{F}_2(\cdot,\bar{\boldsymbol{\theta}})$ is the UNet with learnable parameter $\bar{\boldsymbol{\theta}},\bar{T}$ denotes the number of reverse diffusion steps for the fine-tuning module. In the fine-tuning module, the delivered semantic information is impacted by channel noise, according to \eqref{eq:1}, the semantic information is
\begin{equation} \label{eq:3}
	\mathbf{z}_t=\sqrt{\dot{\alpha_t}}\mathbf{z}_0+\sqrt{1-\dot{\alpha_t}}\boldsymbol{\epsilon}+\boldsymbol{\epsilon}^\mathcal{C}.
\end{equation}
Here, semantic noise $\boldsymbol{\epsilon}^{\mathcal{C}}\sim\mathcal{N}(0,\sigma^2\boldsymbol{I})$ is added after a single transmission, whereas in general diffusion, Gaussian noise is gradually added and denoised through multiple steps. Attempting to denoise all semantic noise within a single step can lead to the loss and distortion of semantic information even when there are only minor noise prediction errors. Therefore, it is essential to simulate the general diffusion process to gradually filter channel-caused semantic noise $\boldsymbol{\epsilon}^{\mathcal{C}}$. If semantic noise is gradually added during transmission, it can likewise be removed step-by-step.

In the modified diffusion process of fine-tuning module, we consider that the semantic noise $\boldsymbol{\epsilon}^{\mathcal{C}}$ is gradually added to image semantic information within steps $\{1,\cdots,t,\cdots,\bar{T}\}$, then the diffusion process can be formulated as
\begin{equation} \label{eq:4}
	\mathbf{z}_t=\sqrt{\dot{\alpha}_t}\mathbf{z}_0+\sqrt{1-\dot{\alpha}_t}\boldsymbol{\epsilon}+\sigma\sum_{t=1}^{\bar{T}}\prod_{j=t+1}^{\bar{T}-1}\sqrt{\gamma_t}\sqrt{\alpha_j}\boldsymbol{\epsilon},
\end{equation}
where $0 < \gamma _1 < \gamma _2 < \cdots < \gamma _t < \cdots < \gamma _{\bar{T} } < 1$ is a variance schedule with time related constants, and $\sum_{t=1}^{\bar{T}}\prod_{j=t+1}^{\bar{T}-1}\sqrt{\gamma_t}\sqrt{\alpha_j}=1$.

Then, the reverse diffusion process from step $t$ to $t-1$ in the fine-tuning module is as
\begin{equation} \label{eq:5}
\begin{aligned}
    	\mathbf{z}_{t-1}=\frac{1}{\sqrt{\alpha_t}}\left(\mathbf{z}_t-C(\alpha,\sigma,t)\boldsymbol{\epsilon}_{\boldsymbol{\theta}}(\mathbf{z}_t,t,\mathbf{z}_{\scriptscriptstyle t\!e\!x\!t})\right)+\bar{\sigma}_t\boldsymbol{\epsilon}, \\
    where\; C(\alpha,\sigma,t)=\frac{\begin{pmatrix}1-\alpha_t+\sigma_{t,t-1}^2\end{pmatrix}\begin{pmatrix}\sqrt{1-\dot{\alpha}_t}-\sigma_t^2\end{pmatrix}}{1-\dot{\alpha}_t+\sigma_{t-1}^2\alpha_t+\sigma_{t,t-1}^2}
\end{aligned}
\end{equation}
is a set of time related constants, and $\boldsymbol{\epsilon}_{\boldsymbol{\theta}}(\mathbf{z}_t,t,\mathbf{z}_{\scriptscriptstyle t\!e\!x\!t})$ is the noise predicted from semantic information $\mathbf{z}_t.$ Additionally, $\sigma_t$, $\sigma_{t-1},\sigma_{t,t-1}$, and $\bar{\sigma}_t$ are time-dependent standard deviations for the random sample of Gaussian noise. Theoretically, the semantic noise $\boldsymbol{\epsilon}^{\mathcal{C}}$ can be denoised according to the modified reverse diffusion process as proofed follows.

The key of the modified reverse diffusion is to predict the distribution probability of the semantic information in the previous step $\mathbf{z}_{t-1}$ with the present information $\mathbf{z}_t$, i.e., $q({\mathbf{z}_{t-1}}|\mathbf{z}_t,\mathbf{z}_0)$. According to the Bayes' theorem, we have
\begin{equation} \label{eq:6}
    \begin{aligned}
    &q(\mathbf{z}_{t-1}|\mathbf{z}_t,\mathbf{z}_0)=q(\mathbf{z}_t|\mathbf{z}_{t-1},\mathbf{z}_0)\frac{q(\mathbf{z}_{t-1}|\mathbf{z}_0)}{q(\mathbf{z}_t|\mathbf{z}_0)},\\
	where~&q(\mathbf{z}_t|\mathbf{z}_0)\sim\mathcal{N}(\sqrt{\dot{\alpha}_t}\mathbf{z}_0,1-\dot{\alpha}_t+\sigma_t^2), \\
	&q(\mathbf{z}_{t-1}|\mathbf{z}_0)\sim\mathcal{N}(\sqrt{\dot{\alpha}_{t-1}}\mathbf{z}_0,1-\dot{\alpha}_{t-1}+\sigma_{t-1}^2), \\
	&q(\mathbf{z}_t|\mathbf{z}_{t-1},\mathbf{z}_0)\sim\mathcal{N}(\sqrt{\dot{\alpha}_t}\mathbf{z}_{t-1},1-\alpha_t+\sigma_{t,t-1}^2).
\end{aligned}
\end{equation}
Then, with the Gaussian distribution formula $f(x)=\frac{1}{\sigma\sqrt{2\pi}}exp\left(-\frac{1}{2}\left(\frac{x-\mu}{\sigma}\right)^2\right)$, we have
\begin{equation} \label{eq:7}
	\begin{aligned}
		q(\mathbf{z}_{t-1|\mathbf{z}_t,\mathbf{z}_0})\propto&\frac{1}{\sigma\sqrt{2\pi}}exp\left(-\frac{1}{2}\left(\frac{\left(\mathbf{z}_t-\sqrt{\dot{\alpha}_t}\mathbf{z}_{t-1}\right)^2}{1-\alpha_t+\sigma_{t,t-1}^2}\right.\right.\\
		&+\!\frac{\left(\mathbf{z}_{t-1}\!-\!\sqrt{\dot{\alpha}_{t-1}}\mathbf{z}_0\right)^2}{1\!-\!\dot{\alpha}_{t-1}\!+\!\sigma_{t-1}^2}\!-\!\frac{\left(\mathbf{z}_t\!-\!\sqrt{\dot{\alpha}_t}\mathbf{z}_0\right)^2}{1\!-\!\dot{\alpha}_t\!+\!\sigma_t^2}\Bigg)\Bigg).
	\end{aligned}
\end{equation}
As Gaussian distribution formula can be transformed as $f(x)=\frac{1}{\sigma\sqrt{2\pi}}exp\left(-\frac{1}{2}\left(\frac{1}{\sigma^{2}}x^{2}-\frac{2\mu}{\sigma^{2}}x+\frac{\mu^{2}}{\sigma^{2}}\right)\right)$, we have $\frac\mu{\sigma^2}=\left(\frac{\sqrt{\alpha_t}\mathbf{z}_t}{1-\alpha_t+\sigma_{t,t-1}^2}+\frac{\sqrt{\dot{\alpha}_{t-1}}\mathbf{z}_0}{1-\dot{\alpha}_{t-1}+\sigma_{t-1}^2}\right)$ and $\frac{1}{\sigma^2}=\left ( \frac {\alpha _t}{1- \alpha _t+ \sigma _{t, t- 1}^2}+ \frac 1{1- \dot{\alpha } _{t- 1}+ \sigma _{t- 1}^2}\right)$. Then, we can calculate the mean of the distribution $q(\mathbf{z}_{t-1}|\mathbf{z}_t,\mathbf{z}_0)$ as
\begin{equation} \label{eq:8}
	\begin{aligned}
		\bar{\mu}(\mathbf{z}_t,\mathbf{z}_0)=&\frac{\sqrt{\alpha_t}\left(1-\dot{\alpha}_{t-1}+\sigma_{t-1}^2\right)}{1-\dot{\alpha}_t+\sigma_{t-1}^2\alpha_t+\sigma_{t,t-1}^2}\mathbf{z}_t+\\
		&\frac{\sqrt{\dot{\alpha}_{t-1}}\left(1-\alpha_t+\sigma_{t,t-1}^2\right)}{1-\dot{\alpha}_t+\sigma_{t-1}^2\alpha_t+\sigma_{t,t-1}^2}\mathbf{z}_0.
	\end{aligned}
\end{equation}

\noindent Then, since \eqref{eq:3} can be reversed as $\mathbf{z}_0\quad=\frac{1}{\sqrt{\dot{\alpha}_t}}\left(\mathbf{z}_t-(\sqrt{1-\dot{\alpha}_t}-\sigma_t^2)\boldsymbol{\epsilon}_t\right),\boldsymbol{\epsilon}_t \sim \mathcal{N}(0,\boldsymbol{I})$, thus we have
\begin{equation} \label{eq:9}
	\bar{\mu}(\mathbf{z}_t,t)\!=\!\frac{1}{\sqrt{\alpha_t}}\!\left(\mathbf{z}_t\!-\!\frac{\left(1\!-\!\alpha_t\!+\!\sigma_{t,t-1}^2\right)\left(\sqrt{1\!-\!\dot{\alpha}_t}\!-\!\sigma_t^2\right)}{1\!-\!\dot{\alpha}_t\!+\!\sigma_{t-1}^2\alpha_t\!+\!\sigma_{t,t-1}^2}\boldsymbol{\epsilon}_t\right).
\end{equation}

\noindent Therefore, we have $\mu_{\boldsymbol{\theta}}(\mathbf{z}_t,t)$ as the mean of distribution $q(\mathbf{z}_{t-1}|\mathbf{z}_t,\mathbf{z}_0).$ Using a neural network to estimate $\boldsymbol\epsilon_t$, and according to the reparameterization trick, we have \eqref{eq:5}.

\textit{3) Autoencoder decoder module:} This is the decoder of the VAE model with KL loss~\cite{12}, which paints the final image output using the fused semantic information of text and image. The final image from the autoencoder decoder module is output as $\bar{s} = \mathcal{D} ( \mathbf{z} _{\bar{\boldsymbol{\theta }} }| \mathbf{z} _{\scriptscriptstyle t\!e\!x\!t}; \bar{\boldsymbol{\varphi }} )$, where $\mathcal{D}(\cdot;\bar{\boldsymbol{\varphi}})$ is the autoencoder decoder with learnable parameter $\bar{\boldsymbol{\varphi}}$.

\section{Training Algorithms and Performance Metrics}

Training algorithm and performance evaluation metrics, as two important parts for guiding high-quality wireless AIGC service delivery, are elaborated in this section.

\begin{figure*}[htbp]
	\centering
	\includegraphics[width=0.56\linewidth]{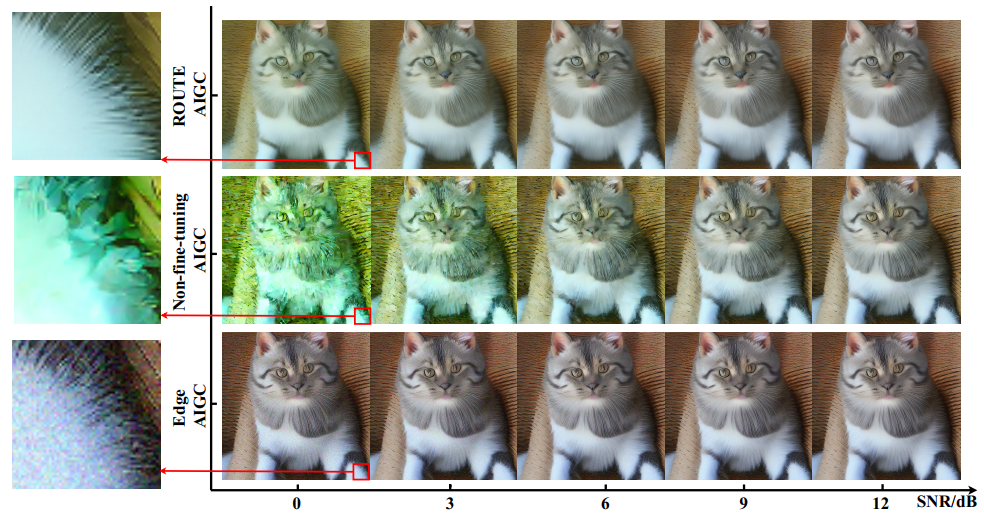}
	\caption{The delivered images of the three frameworks under different SNRs (text input: ``A cute furry cat").}
	\label{fig:3}
\end{figure*}

\begin{figure}[htbp]
	\centering
	\subfloat[Average latency under different percentages of computing resources.]{
		\includegraphics[width=0.6\linewidth]{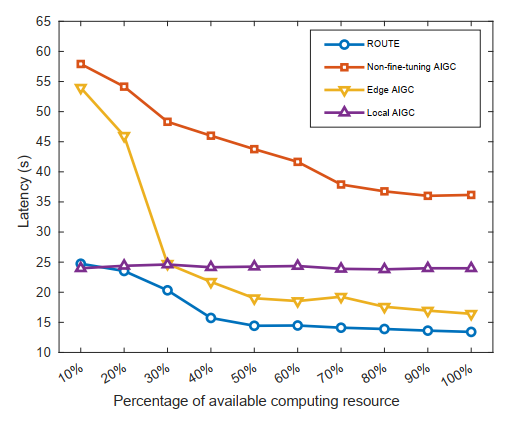}
	} \\
	\subfloat[Average latency under different SNRs.]{
		\includegraphics[width=0.6\linewidth]{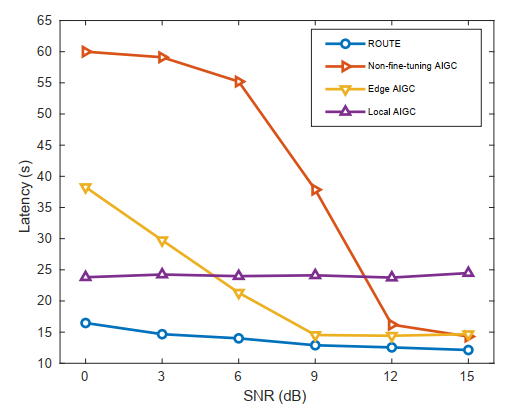}
	}
	\caption{The average latency for fetching one image.}
	\label{fig:4}
\end{figure}

\subsection{Training Algorithms}

In the model training, the image semantic extraction module and the autoencoder decoder module are jointly trained according to the loss function of the VAE encoder-decoder~\cite{12,13} with a dataset $\mathbf{S}=\{\mathbf{s}^{(1)},\cdots,\mathbf{s}^{(i)},\cdots\mathbf{s}^{(M)}\}$, $M$ is the dataset size. The loss function is
\begin{equation} \label{eq:10}
	\begin{aligned}
		loss_1(\boldsymbol{\varphi},\bar{\boldsymbol{\varphi}})=&\frac{1}{2m}\sum_{i=1}^M\sum_{j=1}^K\left(1+\log\sigma_{i,j}^2-\sigma_{i,j}^2-\mu_{i,j}^2\right)\\
		&+\frac{1}{m}\sum_{i=1}^{M}\left(\|\mathbf{s}^{(i)}-\bar{\mathbf{s}}^{(i)}\|_2^2\right),
	\end{aligned}
\end{equation}

\noindent where $K$ is the set size of means $\mu_{i,j}\in\boldsymbol{\mu}_i$ and standard deviations $\sigma_{i,j}\in\boldsymbol{\sigma}_i$ generated for datapoint $\mathbf{s}^{(i)}$ by the VAE encoder. Here, $\sum_{j=1}^{K}\left(1+\log\sigma_{i,j}^{2}-\sigma_{i,j}^{2}-\mu_{i,j}^{2}\right)$ is the semantic similarity term calculated by KL divergence between the approximate Gaussian postier $\mathcal{N}(\boldsymbol{\mu}_i,\boldsymbol{\sigma}_i^2)$ and standard normal ${\mathrm{prior~}\mathcal{N}}(0,\boldsymbol{I}).{\mathrm{~Additionally,~}}\left(\|\mathbf{s}_{i}-\bar{\mathbf{s}}_{i}\|_{2}^{2}\right)$serves as the restructuration likelihood term, measuring the difference between the original image $\bar{\mathbf{s}}_i$ and the reconstructed image $\bar{\mathbf{s}}_i$.

To achieve the efficient generation and transmission of high-quality content, the pre-training for the diffusion-based model of the semantic generation module is the crucial part~\cite{9}.

The corresponding objective of the information generator is to predict the noise distribution in the extracted semantic information as follows
\begin{equation} \label{eq:11}
	loss_2=\mathbb{E}_{\mathcal{E}(s),\boldsymbol{\epsilon}\sim\mathcal{N}(0,\boldsymbol{I}),t}\left[\|\boldsymbol{\epsilon}-\boldsymbol{\epsilon}_{\boldsymbol{\theta}}\left(z,t|\mathbf{z}_{\scriptscriptstyle t\!e\!x\!t}\right)\|_2^2\right].
\end{equation}
For the fine-tuning module, the training objective is similar to \eqref{eq:11}, except its UNet is trained to learn the combined distribution of the semantic noise and additional channel noise-caused semantic noise.

\subsection{T2I Service Evaluation Metrics}
In this study, image quality and latency are the two crucial performance metrics that affect service quality. The image quality is primarily determined by semantic accuracy and construction coherence, and can be evaluated according to an aesthetics scores predictor~\cite{14}. Meanwhile, the latency of a local receiver $j$ accessing T2I service from an edge transmitter $i$ includes transmission delay $L_1$, edge computing delay $L_2$, and local computing delay $L_3$, expressed as $L=L_1+L_2+L_3$. The transmission delay is $L_1=\frac {O}{v_{i,j}}$, where $O$ is the data size of semantic information and $v_{i,j}$ is the bit transmission rate of the link from transmitter $i$ to receiver $j$. In addition, the computing latency $L_2$ and $L_3$ can be directly fetched from the program running times at the transmitter and receiver.

\section{Simulation Results and Discussions}

In this section, we conduct simulations to evaluate the performance of the ROUTE under different scenarios. In order to demonstrate the rationality of combining SemCom and diffusion-based GAI model, we compare the proposed ROUTE with the following three benchmarks.
\begin{enumerate}
	\item Non-fine-tuning AIGC: This framework offers a straightforward application of SemCom to the AIGC transmitter-receiver without a fine-tuning module in the receiver.
	\item Edge AIGC: The image is generated by the T2I GAI model deployed in the edge server and uses traditional communication to deliver to users.
	\item Local AIGC: This framework uses a local receiver to compute all the processes of T2I generation, in which no wireless communication is involved~\cite{9}.
\end{enumerate}

\subsection{Simulation Settings}

In this simulation, an RTX A6000 GPU workstation acts as the edge transmitter, and a GTX1080 GPU laptop functions as the local receiver. The transmitter-to-receiver link has a $20MHz$ bandwidth budget and an SNR range of $[0,15]dB.$ If image generation exceeds $60s$, the task is considered failed. For the semantic generation and fine-tuning models, the \textit{Stable Diffusion v1-5} model checkpoint is used as the pre-trained models~\cite{9}. The corresponding semantic extraction module of text is deployed with pre-trained OpenAI CLIP~\cite{8}. Meanwhile, the pre-trained VAE encoder and decoder are used as the image semantic extraction module and autoencoder decoder module, respectively~\cite{12}. Moreover, the generated image quality is evaluated with the \textit{laion-AI/aesthetic-predictor}. To adjust the UNet weights of the fine-tuning module~\cite{15}, an image-text pair subset within the \textit{laion-aesthetics v2 5+} is used. In training, all the training images are filtered to a size of $512\times512\times3$. All the pre-processed images are extracted as $1\times4\times64\times64$ image semantic information by VAE encoder for UNet training. Moreover, the UNet is trained with 1 middle block in $8\times8$ resolution, 12 encoding blocks, and 12 decoding blocks in 4 resolutions, i.e., $64\times 64$, $32\times 32$, $16\times 16$, and $8\times8$~\cite{9}.

\subsection{Numerical Results}
We first examine the generated image quality under different channel qualities, shown in Fig.~\ref{fig:3}. Moreover, as the number of denoising steps affects image quality, for fairness, we ought to set the total denoising step of different frameworks to the same number of 20~\cite{5}. The ROUTE tends to flexibly adjust the denoising step placed in the transmitter and receiver. Meanwhile, for both the Non-fine-tuning AIGC and Edge AIGC, the denoising steps are solely in the edge transmitter. Since local AIGC does not involve information transmission, it is not set as a benchmark under different SNRs. As shown in Fig.~\ref{fig:3} it is obvious that the images generated by ROUTE exhibit excellent clarity with accurate colors, and align seamlessly with the textual description. The ROUTE outperforms the Non-fine-tuning AIGC framework under all channel quality while generating images with structures that nearly resemble those of the edge AIGC, thanks to the fine-tuning model's effective noise reduction and preservation of valuable semantic information. Moreover, in zoom-in figures, it is observed that the ROUTE can output image with a smooth texture, while images of other frameworks have noisy points and blurred edges. The results further indicate that the proposed ROUTE is effective and robust in generating high-quality images.

Fig.~\ref{fig:4}(a) and Fig.~\ref{fig:4}(b) illustrate the average latency of generating images for four AIGC services under different transmitter computing resources availability and different SNRs. A simple dueling double deep Q network is used to decide the denoising step numbers in ROUTE. It is obvious that under almost all scenarios, the ROUTE can outperform the other three schemes with latency $[12, 25]s$. The superiorities of ROUTE come from its flexibility to adjust computing workload as per dynamic computational resource availability, and it requires fewer communication resources in transmission.

\section{Conclusion}

In this paper, we propose the workload-adjustable transceiver ROUTE for edge-local cooperatively delivering AIGC service in wireless networks. Our framework employs SemCom to reduce the consumption of communication resources. In addition, diffusion models are deployed into both the encoder and decoder to improve the utilization of computational resources, adjust the generated content quality, reduce the service latency. Simulations are conducted under dynamic resource availability and channel quality, where results demonstrate the service quality and robustness improvement of our proposed ROUTE. This work inspired the use of SemCom for cooperative delivery of wireless AIGC services.

\normalem
\bibliographystyle{IEEEtran}
\bibliography{IEEEabrv,bibSemantic}

\end{document}